\documentclass[11pt]{article} 
\usepackage{rldmsubmit,palatino}
\usepackage{graphicx}
\usepackage{times}
\usepackage{soul}
\usepackage{url}
\usepackage[utf8]{inputenc}
\usepackage[small]{caption}
\usepackage{hyperref}
\usepackage{graphicx}
\usepackage{amsmath}
\usepackage{amsthm}
\usepackage{amssymb}
\usepackage{booktabs}
\usepackage{natbib}
\usepackage{algorithm,algpseudocode}
\usepackage{algorithmicx}
\usepackage{breqn}
\usepackage{bm}
\usepackage{comment}
\usepackage{paralist}
\usepackage{multicol}
\usepackage[rightcaption]{sidecap}
\usepackage{subcaption}
\title{A Micro-Objective Perspective of Reinforcement Learning}

\author{
Changjian Li \\
Department of Electrical and Computer Engineering\\
University of Waterloo\\
Waterloo, ON N2L3G1 \\
\texttt{changjian.li@uwaterloo.ca} \\
\And
Krzysztof Czarnecki \\
Department of Electrical and Computer Engineering\\
University of Waterloo\\
Waterloo, ON N2L3G1 \\
\texttt{k2czarne@uwaterloo.ca} \\
}
\DeclareMathOperator*{\argmax}{arg\,max}

\begin{document}

\maketitle

\begin{abstract}
The standard \emph{reinforcement learning} (RL) formulation considers the expectation of the (discounted) cumulative reward. This is limiting in applications where we are concerned with not only the \emph{expected} performance, but also the distribution of the performance. In this paper, we introduce \emph{micro-objective reinforcement learning} --- an alternative RL formalism that overcomes this issue. In this new formulation, a RL task is specified by a set of \emph{micro-objectives}, which are constructs that specify the desirability or undesirability of events. In addition, micro-objectives allow prior knowledge in the form of temporal abstraction to be incorporated into the global RL objective. The generality of this formalism, and its relations to single/multi-objective RL, and hierarchical RL are discussed.
\end{abstract}

\keywords{
reinforcement learning; Markov decision process
}

\acknowledgements{
The authors would like to thank Sean Sedwards, Jaeyoung Lee and other members of Waterloo Intelligent Systems Engineering Lab (WISELab) for discussions.
}

\startmain 

\section{Introduction and Related Works}
The RL formulation commonly adopted in literature aims to maximize the expected \emph{return} (discounted cumulative reward), which is desirable if all we are concerned with is the expectation. However, in many practical problems, especially in risk-sensitive applications, we not only care about the expectation, but also the distribution of the return. For example, in autonomous driving, being able to drive well in expectation is not enough, we need to guarantee that the risk of collision is below a certain acceptable level. As another example, there might be two investment plans with the same expected return, but different variance. Depending on investor type, one investment plan might be more attractive than the other. As a simplified abstraction, we consider the following \emph{Markov Decision Process} (MDP) with only one non-absorbing state $s_{0}$, which is the state where the investment decision is to be made. There are two actions, $a_{1}$ and $a_{2}$, corresponding to the two investment plans. From $s_{0}$, if $a_{1}$ is taken, there is $0.9$ chance of getting a profit of $10$ (entering absorbing state $s_{1}$), and $0.1$ chance of getting a loss of $-10$ (entering absorbing state $s_{2}$). If $a_{2}$ is taken, there is $0.7$ probability of earning a profit of $20$ (entering absorbing state $s_{3}$), and $0.3$ probability of receiving a loss of $-20$ (entering absorbing state $s_{4}$). The reward function is therefore as follows:
\begin{align}
\label{eq:invest_r}
r(s_{0}, a_{1}, s_{1}) = 10, \quad r(s_{0}, a_{1}, s_{2}) = -10, \quad r(s_{0}, a_{2}, s_{3}) = 20, \quad r(s_{0}, a_{2}, s_{4}) = -20
\end{align}
The reward function is zero onwards once an absorbing state is reached. Both $a_{1}$ and $a_{2}$ will result in an expected return of $8$. However, the investor might not be able to afford a loss of more than, say, $15$, in which case $a_{1}$ is preferable to $a_{2}$. Unfortunately, the expected return formulation provides no mechanism to differentiate these two actions. Furthermore, any mixture policy of $a_{1}$ and $a_{2}$ (mixing $a_{1}$ and $a_{2}$ with some probability) also has the same expected return.

Two approaches have been discussed in literature to tackle this issue. One is to shape the reward so that the expected return of the policies are no longer the same~\citep{DBLP:conf/kr/KoenigS94}. E.g., we can give more negative reward when the loss is higher than $15$,  such as the following:
\begin{displaymath}
  r'(s, a, s')= 
  \begin{cases}
    r(s, a, s') ,& \text{if } r(s, a, s') \geq -15\\
    r(s, a, s') - (r(s, a, s') + 15)^2 ,& \text{otherwise}
  \end{cases}
\end{displaymath}
Although in this simple case, the reward shaping is fairly straight-forward, in more complex tasks such as autonomous driving where there are many conflicting aspects, it is a challenge to choose a reward that properly balances the expected overall performance and the risk of each aspect. 

The second approach is to use an alternative formulation that considers more than just the expected return. Several methods~\citep{Sato2001, DBLP:conf/uai/SherstanABYWWS18, DBLP:conf/icml/TamarCM13} have been proposed to estimate the variance of return in addition to the expectation. While this alleviates the issue by taking variance into account, distributions can differ even if both the expectation and the variance are the same. \citeauthor{YU1998193}~\citep{YU1998193} considered the problem of maximizing the probability of receiving a return that is greater than a certain threshold. \citeauthor{DBLP:journals/corr/abs-1109-2147}~\citep{DBLP:journals/corr/abs-1109-2147} considered constrained MDPs~\citep{cMDP} where the discounted probabilities of error states (unacceptable states) are constrained. However, both of these formulations are designed only for a specific type of application, and do not have the generality required as an alternative RL formalism.

In this paper, we propose to solve this issue by restricting the return to a distribution that is entirely decided by its mean, namely the Bernoulli distribution. To motivate this idea, observe that in any RL task, we are essentially concerned with a set of events, some desirable, some undesirable, to different extents. For example, in the task of autonomous driving, collision is an undesirable event; running a red light is another event, still undesirable but not as much; driving within the speed limit is a desirable event, etc. Instead of associating each event with a reward, and evaluating a policy by the total reward it accumulates (as in conventional RL), we can think of all the events as a whole, and evaluate a policy based on the combination of events it would lead to. The return is now only an indicator of an event, and is restricted to binary values: $1$ if the event happens, and $0$ if it does not. Given a policy, the return of each micro-objective is thus a Bernoulli random variable, whose mean is both the \emph{value function}, and the probability of event occurrence under the policy. Following this view, a task can be specified by a predefined set of events, and a partial order that allows comparison between different combinations of event probabilities. The goal is to find the policies that result in the most desirable combinations of probabilities. \footnote{To be exact, the combinations (of probabilities) that are not less desirable than any other combinations.}.

This can be illustrated with the investment example. Instead of defining a reward function as in Eq. \ref{eq:invest_r}, we define entering $s_{1}$, $s_{2}$, $s_{3}$ and $s_{4}$ as four events. If action $a_{1}$ is taken, the chances of the four events occurring are $0.9$, $0.1$, $0$ and $0$. If action $a_{2}$ is taken, the chances are $[0, 0, 0.7, 0.3]$. Apart from the events, a partial order on the probability vectors is also defined to specify which probability vector is more desirable. Let $\mathbf{v}^{\pi} = [v^{\pi}_{1}, v^{\pi}_{2}, v^{\pi}_{3}, v^{\pi}_{4}]$ denote the expected returns of the four events. If we set the partial order to $\mathbf{v}^{\pi} \preceq \mathbf{v}^{\pi'} \iff  20(v^{\pi}_{3} - v^{\pi}_{4}) + 10(v^{\pi}_{1} - v^{\pi}_{2}) \leq 20(v^{\pi'}_{3} - v^{\pi'}_{4}) + 10(v^{\pi'}_{1} - v^{\pi'}_{2})$, we arrive at an equivalent formulation to the standard RL formulation with reward specified by Eq. \ref{eq:invest_r}. If, however, we want the probability of getting a loss of more than $15$ to be less than a certain threshold $\epsilon$, we can simply redefine the partial order so that $\mathbf{v}^{\pi}$ is smaller whenever $v^{\pi}_{4} \geq \epsilon$.

\section{Background}
\subsection{MDP and Reinforcement Learning}
The standard RL problem is often formulated in terms of a (single-objective) \emph{Markov Decision Process} (MDP), which can be represented by a six-tuple $(S, A, P, \mu, \gamma, r)$, where $S$ is a finite set of states; $A$ is a finite set of actions; $P(s'|s, a)$ is the transition probability from state $s$ to state $s'$ taking action $a$; $\mu$ is the initial state distribution; $\gamma \in [0, 1]_{\mathbb{R}}$ is the discount factor; and $r(s, a, s') \in \mathbb{R}$ is the reward for taking action $a$ in state $s$ and arriving state $s'$. The goal is to find the maximal expected discounted cumulative reward: $\max_{\pi \in \Pi} \mathbb{E}[\sum_{t=0}^{\infty}{\gamma^{(t)}r(s^{t}, a^{t}, s^{t+1})}| \pi, \mu]$, where $\gamma^{(t)}$ denotes $\gamma$ to the power of $t$, and $\Pi$ denotes the set of policies we would like to consider. In the most general case, a policy can be history-dependent and random, in the form of $\pi = (\pi^{0}, \pi^{1}, ..., \pi^{t}, ...)$, where a \emph{decision rule} $\pi^{t}(h^{t}, s, a) \in [0, 1]_{\mathbb{R}}$ is the probability of taking action $a$ in state $s$ with history $h^{t}$. A history is a sequence of past states, actions and decision rules $h^{t} = (s^{0}, a^{0}, \pi^{0}, s^{1}, a^{1}, \pi^{1}, ..., s^{t-1}, a^{t-1}, \pi^{t-1})$. A policy is said be \emph{deterministic} if $\pi^{t}(h^{t}, s, a) = 1$ for only one action, in which case we can use a simplified notation $\pi = (d^{0}, d^{1}, ..., d^{t}, ...)$, where $d^{t}(h^{t}, s) \in A$. Correspondingly, if the policy is deterministic, a history can be represented with $h^{t} = (s^{0}, d^{0}, s^{1}, d^{1}, ..., s^{t-1}, d^{t-1})$.  A policy is said to be \emph{stationary} if the decision rule only depends on the current state $s$, and does not change with time, i.e., $\pi^{t}(h^{t}, s, a) = \pi(s, a)$. The set of all history-dependent random policies, history-dependent deterministic policies, stationary random policies, and stationary deterministic policies are denoted by $\Pi_{\text{HR}}$, $\Pi_{\text{HD}}$, $\Pi_{\text{SR}}$ and $\Pi_{\text{SD}}$, respectively. We call a task an \emph{episodic} task with \emph{horizon} $T$ if the state space is augmented with time; $\gamma=1$; and $r(s, a, s')=0,  \forall t \geq T$. The expected return following a policy starting from the initial state distribution is called the value function, which is denoted as $v^{\pi}$. For a single-objective MDP, there exists stationary deterministic optimal policy, that is, $\exists \pi \in \Pi_{\text{SD}}, v^{\pi} = \argmax_{\pi' \in \Pi_{\text{HR}}} v^{\pi'}$

\subsection{Multi-objective Reinforcement Learning} \label{subsec: MORL}
In some cases~\citep{DBLP:journals/corr/RoijersVWD14}, it is preferable to consider different aspects of a task as separate objectives. \emph{Multi-objective reinforcement learning} is concerned with \emph{multi-objective Markov decision processes} (MOMDPs) $(S, A, P, \mu, [(\gamma_{1}, r_{1}), ..., (\gamma_{i}, r_{i}), ..., (\gamma_{k}, r_{k})], \preceq)$ , where $S$, $A$ and $P(s'|s, a)$, $\mu$ are the state space, action space,  transition probability and initial state distribution as in single-objective MDPs; Now there are k pairs of discount factors $\gamma_{i}$ and rewards $r_{i}(s, a, s')$, one for each objective. The value function for the $i$\textsuperscript{th} objective is defined as $v_{i}^{\pi} = \mathbb{E}[\sum_{t=0}^{\infty}{\gamma_{i}^{(t)}r_{i}(s^{t}, a^{t}, s^{t+1})} | \pi, \mu]$. Let $\mathbf{v}^{\pi} = [v^{\pi}_{1}, v^{\pi}_{2}, ..., v^{\pi}_{k}]$ be the value functions for all objectives, and $V^{\Pi} = \{\mathbf{v}^{\pi} | \pi \in \Pi\}$ be the set of all realizable value functions by policies in $\Pi$, $\preceq$ is a partial order defined on $V^{\Pi_{\text{HR}}}$. Multi-objective RL aims to find the policies $\pi \in \Pi$ such that $\mathbf{v}^{\pi}$ is a \emph{maximal element}~\footnote{A maximal element of a subset $X$ of some partially ordered set is an element of $X$ that is not smaller than any other element in $X$.} of $V^{\Pi}$, which we refer to as the \emph{optimal} policies. We say a policy $\pi$ \emph{strictly dominates} policy $\pi'$ if $\mathbf{v}^{\pi'} \preceq \mathbf{v}^{\pi}$ and $\mathbf{v}^{\pi} \not\preceq \mathbf{v}^{\pi'}$. A commonly adopted partial order for multi-objective RL is $\mathbf{v}^{\pi} \preceq \mathbf{v}^{\pi'} \iff \forall i \in \{1, 2, ..., k\}, v^{\pi}_{i} \leq v^{\pi'}_{i}$, in which case, the set of maximal elements are also called the \emph{Pareto frontier} of $V^{\Pi}$. Episodic tasks have not been widely discussed in the context of multi-objective RL, and most existing literature assumes $\gamma_{1} = \gamma_{2} = ... = \gamma_{k}$. Although for a single-objective MDP, the optimal value can be attained by a deterministic stationary policy, this is in general not true for multi-objective MDPs. \citeauthor{WHITE1982639}~\citep{WHITE1982639} showed that history-dependent deterministic policies can \emph{Pareto dominate} stationary deterministic policies; \citeauthor{DBLP:conf/stacs/ChatterjeeMH06}~\citep{DBLP:conf/stacs/ChatterjeeMH06} proved for the case $\gamma_{1} = \gamma_{2} = ... = \gamma_{k}$ that stationary random policies suffice for \emph{Pareto optimality}.

\section{Micro-Objective Reinforcement Learning}
In standard multi-objective RL, there is no restriction on the reward function of each objective. Each objective can itself be a `macro' objective that involves multiple aspects. This makes multi-objective RL subject to the same issue single-objective RL has: only the \emph{expectation} of return is considered for each objective. Conceptually, micro-objective RL is multi-objective RL at its extreme: each \emph{micro-objective} is concerned with one and only one aspect --- the occurrence of an \emph{event}. An event can be `entering a set of goal/error states', `taking certain actions in certain states', or `entering a set of states at certain time steps', etc., but ultimately can be represented by a set of histories. If the history up to the current time step $(h^{t}, s^{t})$ is in the set, we say that the event happens.

\subsection{Micro-objectives}
At the core of the micro-objective formulation is a new form of value function $v^{\pi}_{\psi_{i}, T_{i}}(\phi_{i} | \mu)$. Denoting $H$ as the set of all possible histories, $\psi_{i} \subset H$ is the set of histories that corresponds to the occurrence of the event, which we call the \emph{termination set} of a micro-objective. $\phi_{i} \subset H$ is also a set of histories, which we call the \emph{initiation set}. The terminologies are deliberately chosen to resemble those of \emph{options}~\citep{DBLP:journals/ai/SuttonPS99}, and as we will see, this form of value function is indeed connected to options. Independent from the task, a micro-objective has its own initiation and termination. A micro-objective initiates if it is not currently active and $(h^{t}, s^{t}) \in \phi_{i}$, when an associated timer $t_{i}$ is also initiated. A micro-objective terminates if it is currently active and $(h^{t}, s^{t}) \in \psi_{i}$, upon which a return of $1$ is received. It also terminates if $t_{i} \geq T_{i}$ or the task terminates, upon which a return of $0$ is received, and $t_{i}$ is reset. Note that $t$ and $T$ are the time step and time horizon for the task, whereas $t_{i}$ and $T_{i}$ are the time step and time horizon for the micro-objective. $v^{\pi}_{\psi_{i}, T_{i}}(\phi_{i} | \mu)$ is defined as the expected return the $i$\textsuperscript{th} micro-objective receives starting from initial state distribution $\mu$ following policy $\pi$. For example, suppose that the task always starts from $s_{0}$ (i.e., $\mu(s_{0}) = 1$), and the micro-objective is active three times (in sequence) before task termination if policy $\pi$ is followed, with the return of $0$, $0$ and $1$, respectively, then $v^{\pi}_{\psi_{i}, T_{i}}(\phi_{i} | \mu)$ is $\frac{1}{3}(0 + 0 + 1) = \frac{1}{3}$. Although this particular form of value function is similar to the generalized value function (GVF) proposed by \citeauthor{DBLP:conf/atal/SuttonMDDPWP11}~\citep{DBLP:conf/atal/SuttonMDDPWP11} in the sense that both have their own initiation, termination and return, it is a rather different concept. Unlike a GVF which is associated with a \emph{target policy}, and can be interpreted as the answer to a question regarding the target policy; the value function of a micro-objective is parameterized by the global control policy, and is an evaluation of the global policy with respect to one aspect of the task. This becomes clearer when we consider the fact that the value function for a micro-objective is conditioned on the initial state distribution of the task. As a result, the value functions of the micro-objectives appear in the global RL objective, while it is not obvious how GVFs can be used in the RL task specification.

Formally, a micro-objective RL problem is \emph{an episodic task} represented by the $8$-tuple $(S, A, P, \mu, \psi, T, \allowbreak [(\phi_{1}, \psi_{1}, T_{1}), ...,\allowbreak (\phi_{i}, \psi_{i}, T_{i}),\allowbreak ..., (\phi_{k}, \psi_{k}, T_{k})], \preceq)$, where $S$, $A$, $P$, $\mu$ are the state space, action space, transition probabilities, and initial state distribution as usual. $\psi \subset S$ is the terminal states of the task, and $T$ is the time horizon. The task terminates whenever $t \geq T$, or a state in $\psi$ is reached. $(\phi_{1}, \psi_{1}, T_{1})$ to $(\phi_{k}, \psi_{k}, T_{k})$ are the $k$ micro-objectives as described above. Let $\mathbf{v}^{\pi} = [v^{\pi}_{\psi_{1}, T_{1}}(\phi_{1} | \mu), v^{\pi}_{\psi_{2}, T_{2}}(\phi_{2} | \mu), ..., v^{\pi}_{\psi_{k}, T_{k}}(\phi_{k} | \mu)]$ be the value functions for all micro-objectives, and $V^{\Pi} = \{\mathbf{v}^{\pi} | \pi \in \Pi\}$ be the set of all realizable value functions by policies in $\Pi$, $\preceq$ is a partial order defined on $V^{\Pi_{\text{HR}}}$. Similar to multi-objective RL, the goal is to find the policies $\pi \in \Pi$ such that $\mathbf{v}^{\pi}$ is a maximal element of $V^{\Pi}$. However, the multi-objective RL formulation introduced in Section \ref{subsec: MORL} does not subsume micro-objective RL. For one thing, there is no notion of objective termination in multi-objective RL.

\subsection{Connections to Hierarchical RL}
Hierarchical RL~\citep{DBLP:journals/deds/BartoM03} refers to RL paradigms that exploits \emph{temporal abstraction} to facilitate learning, where a higher level policy selects `macro' actions that in turn `call' some lower level actions. Notable Hierarchical RL approaches include the \emph{options} formalism~\citep{DBLP:journals/ai/SuttonPS99}, \emph{hierachical abstract machines} (HAM)~\citep{DBLP:conf/nips/ParrR97}, the MAXQ framework~\citep{DBLP:journals/jair/Dietterich00}, and \emph{feudal RL}~\citep{DBLP:conf/nips/DayanH92}. However, in none of these frameworks does the specification for temporal abstraction appear in the global RL objective. As a result, there is no clear measure on \emph{how well} each temporally abstracted action should be learned, and \emph{how important} they are compared to the goal of the high-level task. The micro-objective formulation is designed to allow temporal abstraction to be expressed as micro-objectives, therefore building the bridge between `objectives' and `options'. To see this, recall that a micro-objective has an initiation set $\phi_{i}$ and a termination set $\psi_{i}$, corresponding to the initiation set $I \subset S$ and termination condition $\beta: S \to [0, 1]_{\mathbb{R}}$ of an option. If the initiation set $I$ and the goal states~\footnote{Options do not need to have goal states. This is an intentional oversimplification to illustrate the idea.} of an option coincide with the initiation set $\phi_{i}$ and the termination set $\psi_{i}$ of a micro-objective, then the micro-objective can be thought of as a measure of how important this option is.
\begin{SCfigure}[1.5][h]
\caption{A task with three hypothetical options. $s_{0}$ is the initial state, and $s_{1}$ is the goal state. There are two possible scenarios. In the first scenario, reaching $\psi$ is important, in which case both $o_{1}$ and $o_{2}$ must be learned well; In the second scenario, $\psi$ is just some guidance for exploration, in which case only $o_{0}$ needs to be learned well. The importance of learning each option can be represented by micro-objectives.}
\includegraphics[scale=0.6]{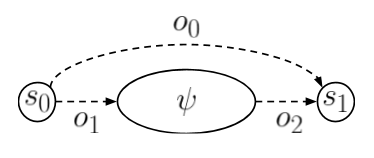}
\end{SCfigure}
To be more concrete, consider the following task where the initial state is $s_{0}$, the goal state is $s_{1}$, and $\psi$ is a set  of intermediate states. $o_{0}$ is a hypothetical (might not exist, and is to be learned if needed) option that takes the agent from $s_{0}$ to $s_{1}$ (might or might not pass through $\psi$). Similarly, $o_{1}$ is a hypothetical option from $s_{0}$ to $\psi$, and $o_{2}$ is a hypothetical option from $\psi$ to $s_{1}$. Such a task can be a taxi agent at location $s_{0}$ driving a passenger from a pick-up location $\psi$ to a destination $s_{1}$, in which case we can define two micro-objectives $(\{s_{0}\}, \psi, T_{1})$ and $(\psi, \{s_{1}\}, T_{2})$, corresponding to $o_{1}$ and $o_{2}$ respectively. In this example, a micro-objective for $o_{0}$ is not needed, because passing though $\psi$ is required. The value function is thus $\mathbf{v}^{\pi} = [v^{\pi}_{\psi, T_{1}}(\{s_{0}\} | \mu(s_{0})=1), v^{\pi}_{\{s_{1}\}, T_{2}}(\psi | \mu(s_{0})=1)]$, and the partial order can be defined as $\mathbf{v}^{\pi} \preceq \mathbf{v}^{\pi'} \iff v^{\pi}_{\psi, T_{1}}(\{s_{0}\} | \mu(s_{0})=1) \leq v^{\pi'}_{\psi, T_{1}}(\{s_{0}\} | \mu(s_{0})=1) \land v^{\pi}_{\{s_{1}\}, T_{2}}(\psi | \mu(s_{0})=1) \leq v^{\pi'}_{\{s_{1}\}, T_{2}}(\psi | \mu(s_{0})=1)$. Another possible scenario for such a task would be an agent navigating through a maze, and $\psi$ is only some heuristics. In this case, the only important micro-objective is $(\{s_{0}\}, \{s_{1}\}, T_{3})$, which corresponds to option $o_{3}$. $o_{1}$ and $o_{2}$ are only there to help exploration. Let $\mathbf{v}^{\pi} = [v^{\pi}_{\psi, T_{1}}(\{s_{0}\} | \mu(s_{0})=1), v^{\pi}_{\{s_{1}\}, T_{2}}(\psi | \mu(s_{0})=1), v^{\pi}_{\{s_{1}\}, T_{3}}(\{s_{0}\} | \mu(s_{0})=1)]$, the partial order can be defined as
\begin{displaymath}
\begin{split}
\mathbf{v}^{\pi} \preceq \mathbf{v}^{\pi'} \iff & v^{\pi}_{\{s_{1}\}, T_{3}}(\{s_{0}\} | \mu(s_{0})=1) < v^{\pi'}_{\{s_{1}\}, T_{3}}(\{s_{0}\} | \mu(s_{0})=1) \lor \bigg( v^{\pi}_{\{s_{1}\}, T_{3}}(\{s_{0}\} | \mu(s_{0})=1) = v^{\pi'}_{\{s_{1}\}, T_{3}}(\{s_{0}\} | \mu(s_{0})=1) \land \\
& v^{\pi}_{\psi, T_{1}}(\{s_{0}\} | \mu(s_{0})=1) \leq v^{\pi'}_{\psi, T_{1}}(\{s_{0}\} | \mu(s_{0})=1) \land v^{\pi}_{\{s_{1}\}, T_{2}}(\psi | \mu(s_{0})=1) \leq v^{\pi'}_{\{s_{1}\}, T_{2}}(\psi | \mu(s_{0})=1) \bigg)
\end{split}
\end{displaymath}

\subsection{Generality}
We briefly discuss the generality of the micro-objective RL formulation. Since the state space, action space and transition probabilities are the same as in a MDP, our only concern is whether micro-objectives, together with the partial order can imply an arbitrary optimal policy. If for any stationary deterministic policy $\pi^{*}$ and any Markov dynamics $(S, A, P)$, there exists a partial order and a set of micro-objectives such that $\pi^{*}$ is optimal, then we can cast any single-objective RL problem into an equivalent micro-objective problem. Now we show that micro-objective RL can indeed imply an arbitrary stationary deterministic policy $\pi^{*} \in \Pi_{\text{SD}}$. Let $s_{1}, s_{2}, ... s_{|S|}$ be an enumeration of the finite state space $S$, we define $|S|$ micro-objectives $v^{\pi}_{\{(h, s_{j}, \pi^{*}(s_{j})) | h \in H\}, 1}(\{(h, s_{j}) | h \in H\} | \mu), j=1, 2, ..., |S|$, which, with abuse of notation, we write as $v^{\pi}_{j}$. In other words, for each state $s \in S$ we define an event: taking $\pi^{*}(s)$ in $s$. If we further define the partial order as $\mathbf{v}^{\pi} \preceq \mathbf{v}^{\pi'} \iff v^{\pi}_{1} \leq v^{\pi'}_{1} \land ... \land v^{\pi}_{|S|} \leq v^{\pi'}_{|S|}$, then $\pi^{*}$ is optimal for this micro-objective RL task. Therefore, given enough micro-objectives, the micro-objective formulation is at least as general as the standard RL formulation.

Although we've shown above that micro-objective RL is able to imply any stationary deterministic policy, it remains interesting whether and how an `objective' in single/multi-objective RL can be directly translated to one or more `micro-objective(s)'. Consider the $i$\textsuperscript{th} objective of multi-objective RL (in the case of single-objective RL, $i=1$) with reward $r_{i}(s, a, s')$. If we define the following (countably infinite) set of micro-objectives:
\begin{displaymath}
\bigg\{ \bigg( \{(h^{t}, s) | h^{t} \in H^{t} \}, \{(h^{t}, s, a, s') | h^{t} \in H^{t} \}, t \bigg) \bigg| s, s' \in S, , t \in \{1, 2, ...\} \bigg\}
\end{displaymath}
where $H^{t}$ denotes the set of all possible histories until (but excluding) time $t$, then the value function $v^{\pi}_{i}$ for the $i$\textsuperscript{th} objective of the original single/multi-objective RL can be written in terms of the above micro-objectives as:
\begin{displaymath}
v^{\pi}_{i} = \sum_{t=0}^{t=\infty} \sum_{s \in S} \sum_{a \in A} \sum_{s' \in S} \gamma^{(t)} r_{i}(s, a, s') v^{\pi}_{\{(h^{t}, s, a, s') | h^{t} \in H^{t} \}, t}(\{(h^{t}, s) | h^{t} \in H^{t} \} | \mu)
\end{displaymath}
thus casting an `objective' as a set of `micro-objectives'.

\subsection{Optimal Policies}
\begin{SCfigure}[0.4][h]
    \centering
    \begin{subfigure}[b]{0.2\textwidth}
    \includegraphics[width=\textwidth]{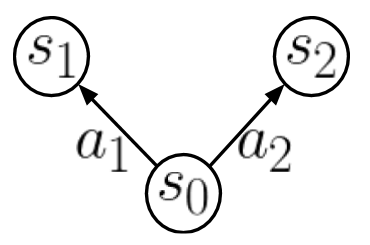}
    \caption{}
    \label{fig:opt1}
    \end{subfigure}
    ~\qquad\quad
    \begin{subfigure}[b]{0.3\textwidth}
    \includegraphics[width=\textwidth]{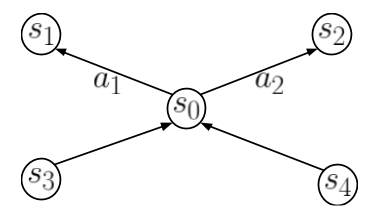}
    \caption{}
    \label{fig:opt2}
    \end{subfigure}
    \caption{The optimal policies for micro-objective RL are in general history-dependent and random.}
\end{SCfigure}

In this section, we discuss the optimal policies for micro-objective RL. We first show that stationary random policies can strictly dominate stationary deterministic policies. Consider the following example (Figure. \ref{fig:opt1}) where there are three states $s_{0}$, $s_{1}$ and $s_{2}$. From $s_{0}$ there are two actions $a_{1}$ and $a_{2}$. $a_{1}$ always leads to state $s_{1}$ and $a_{2}$ always leads to state $s_{2}$. The episode always starts from $s_{0}$, i.e., $\mu(s_{0}) = 1$. There are two micro-objectives $(\{s_{0}\}, \{(h, s_{1}) | h \in H\}, 1)$ and $(\{s_{0}\}, \{(h, s_{2}) | h \in H\}, 1)$, whose value functions are denoted with abuse of notation, as $v^{\pi}_{1}$ and $v^{\pi}_{2}$, respectively. The partial order $\preceq$ is defined as $\mathbf{v}^{\pi} \preceq \mathbf{v}^{\pi'} \iff v^{\pi}_{1}v^{\pi}_{2} \leq v^{\pi'}_{1}v^{\pi'}_{2}$. For this task, the stationary random policy that selects $a_{1}$ and $a_{2}$ with equal probability strictly dominates any stationary deterministic policy that always selects either $a_{1}$ or $a_{2}$.

Now we show that history-dependent deterministic policies can strictly dominate stationary policies. Consider the example (Figure. \ref{fig:opt2}) where there are five states, $s_{0}$ to $s_{4}$. The episode starts from $s_{3}$ and $s_{4}$ with equal probability, i.e., $\mu(s_{3})=\mu(s_{4})=0.5$. The next state for both $s_{3}$ and $s_{4}$ is always $s_{0}$ regardless of the action taken. From $s_{0}$ there are two actions $a_{1}$ and $a_{2}$, leading to $s_{1}$ and $s_{2}$ with certainty, respectively. There are two micro-objectives: to go from $s_{3}$ to $s_{2}$, i.e., $(\{s_{3}\}, \{(h, s_{2}) | h \in H\}, 2)$; and to go from $s_{4}$ to $s_{1}$, i.e., $(\{s_{4}\}, \{(h, s_{1}) | h \in H\}, 2)$. Again we abbreviate the value function of the two micro-objectives respectively as $v^{\pi}_{1}$ and $v^{\pi}_{2}$. The partial order is thus defined as $\mathbf{v}^{\pi} \preceq \mathbf{v}^{\pi'} \iff v^{\pi}_{1} \leq v^{\pi'}_{1} \land v^{\pi}_{2} \leq v^{\pi'}_{2}$. In this task, the only distinction between policies is the choice of action in state $s_{0}$. It's not hard to see that the history-dependent policy that takes action $a_{2}$ in $s_{0}$ if the initial state is $s_{3}$, and takes action $a_{1}$ in $s_{0}$ if the initial state is $s_{4}$ strictly dominates any stationary policies. 

By combining the above two examples, we can construct an example where a history-dependent random policy strictly dominates history-dependent deterministic policies. Therefore the optimal policies for micro-objective RL are in general history-dependent and random.

\section{Conclusions}
We introduced micro-objective RL, a general RL formalism that not only solves the problem of standard RL formulation that only the expectation is considered, but also allows temporal abstraction to be incorporated into the global RL objective. Intuitively, this micro-objective paradigm bears more resemblance to how humans perceive a task --- it is hard for a human to tell what reward they receive at a certain time, but it is relatively easy to tell how good a particular combination of events is. Ongoing research topics include effective algorithms for micro-objective RL, and compact representation of similar micro-objectives.

\bibliographystyle{plainnat}
\bibliography{rldm}

\begin{thebibliography}{16}
\providecommand{\natexlab}[1]{#1}
\providecommand{\url}[1]{\texttt{#1}}
\expandafter\ifx\csname urlstyle\endcsname\relax
  \providecommand{\doi}[1]{doi: #1}\else
  \providecommand{\doi}{doi: \begingroup \urlstyle{rm}\Url}\fi

\bibitem[Altman(1999)]{cMDP}
Eitan Altman.
\newblock \emph{Constrained Markov Decision Processes}.
\newblock Chapman \& Hall/CRC, 1999.
\newblock ISBN 9780849303821.

\bibitem[Barto and Mahadevan(2003)]{DBLP:journals/deds/BartoM03}
Andrew~G. Barto and Sridhar Mahadevan.
\newblock Recent advances in hierarchical reinforcement learning.
\newblock \emph{Discrete Event Dynamic Systems}, 13\penalty0 (1-2):\penalty0
  41--77, 2003.
\newblock \doi{10.1023/A:1022140919877}.
\newblock URL \url{https://doi.org/10.1023/A:1022140919877}.

\bibitem[Chatterjee et~al.(2006)Chatterjee, Majumdar, and
  Henzinger]{DBLP:conf/stacs/ChatterjeeMH06}
Krishnendu Chatterjee, Rupak Majumdar, and Thomas~A. Henzinger.
\newblock Markov decision processes with multiple objectives.
\newblock In \emph{{STACS} 2006, 23rd Annual Symposium on Theoretical Aspects
  of Computer Science, Marseille, France, February 23-25, 2006, Proceedings},
  pages 325--336, 2006.
\newblock \doi{10.1007/11672142\_26}.
\newblock URL \url{https://doi.org/10.1007/11672142\_26}.

\bibitem[Dayan and Hinton(1992)]{DBLP:conf/nips/DayanH92}
Peter Dayan and Geoffrey~E. Hinton.
\newblock Feudal reinforcement learning.
\newblock In \emph{Advances in Neural Information Processing Systems 5, {[NIPS}
  Conference, Denver, Colorado, USA, November 30 - December 3, 1992]}, pages
  271--278, 1992.
\newblock URL
  \url{http://papers.nips.cc/paper/714-feudal-reinforcement-learning}.

\bibitem[Dietterich(2000)]{DBLP:journals/jair/Dietterich00}
Thomas~G. Dietterich.
\newblock Hierarchical reinforcement learning with the {MAXQ} value function
  decomposition.
\newblock \emph{J. Artif. Intell. Res.}, 13:\penalty0 227--303, 2000.
\newblock \doi{10.1613/jair.639}.
\newblock URL \url{https://doi.org/10.1613/jair.639}.

\bibitem[Geibel and Wysotzki(2011)]{DBLP:journals/corr/abs-1109-2147}
Peter Geibel and Fritz Wysotzki.
\newblock Risk-sensitive reinforcement learning applied to control under
  constraints.
\newblock \emph{CoRR}, abs/1109.2147, 2011.
\newblock URL \url{http://arxiv.org/abs/1109.2147}.

\bibitem[Koenig and Simmons(1994)]{DBLP:conf/kr/KoenigS94}
Sven Koenig and Reid~G. Simmons.
\newblock Risk-sensitive planning with probabilistic decision graphs.
\newblock In \emph{Proceedings of the 4th International Conference on
  Principles of Knowledge Representation and Reasoning (KR'94). Bonn, Germany,
  May 24-27, 1994.}, pages 363--373, 1994.

\bibitem[Parr and Russell(1997)]{DBLP:conf/nips/ParrR97}
Ronald Parr and Stuart~J. Russell.
\newblock Reinforcement learning with hierarchies of machines.
\newblock In \emph{Advances in Neural Information Processing Systems 10,
  {[NIPS} Conference, Denver, Colorado, USA, 1997]}, pages 1043--1049, 1997.
\newblock URL
  \url{http://papers.nips.cc/paper/1384-reinforcement-learning-with-hierarchies-of-machines}.

\bibitem[Roijers et~al.(2014)Roijers, Vamplew, Whiteson, and
  Dazeley]{DBLP:journals/corr/RoijersVWD14}
Diederik~Marijn Roijers, Peter Vamplew, Shimon Whiteson, and Richard Dazeley.
\newblock A survey of multi-objective sequential decision-making.
\newblock \emph{CoRR}, abs/1402.0590, 2014.
\newblock URL \url{http://arxiv.org/abs/1402.0590}.

\bibitem[Sato et~al.(2001)Sato, Kimura, and Kobayashi]{Sato2001}
Makoto Sato, Hajime Kimura, and Shibenobu Kobayashi.
\newblock Td algorithm for the variance of return and mean-variance
  reinforcement learning.
\newblock \emph{Transactions of the Japanese Society for Artificial
  Intelligence}, 16\penalty0 (3):\penalty0 353--362, 2001.
\newblock \doi{10.1527/tjsai.16.353}.

\bibitem[Sherstan et~al.(2018)Sherstan, Ashley, Bennett, Young, White, White,
  and Sutton]{DBLP:conf/uai/SherstanABYWWS18}
Craig Sherstan, Dylan~R. Ashley, Brendan Bennett, Kenny Young, Adam White,
  Martha White, and Richard~S. Sutton.
\newblock Comparing direct and indirect temporal-difference methods for
  estimating the variance of the return.
\newblock In \emph{Proceedings of the Thirty-Fourth Conference on Uncertainty
  in Artificial Intelligence, {UAI} 2018, Monterey, California, USA, August
  6-10, 2018}, pages 63--72, 2018.
\newblock URL \url{http://auai.org/uai2018/proceedings/papers/35.pdf}.

\bibitem[Sutton et~al.(1999)Sutton, Precup, and
  Singh]{DBLP:journals/ai/SuttonPS99}
Richard~S. Sutton, Doina Precup, and Satinder~P. Singh.
\newblock Between mdps and semi-mdps: {A} framework for temporal abstraction in
  reinforcement learning.
\newblock \emph{Artif. Intell.}, 112\penalty0 (1-2):\penalty0 181--211, 1999.
\newblock \doi{10.1016/S0004-3702(99)00052-1}.
\newblock URL \url{https://doi.org/10.1016/S0004-3702(99)00052-1}.

\bibitem[Sutton et~al.(2011)Sutton, Modayil, Delp, Degris, Pilarski, White, and
  Precup]{DBLP:conf/atal/SuttonMDDPWP11}
Richard~S. Sutton, Joseph Modayil, Michael Delp, Thomas Degris, Patrick~M.
  Pilarski, Adam White, and Doina Precup.
\newblock Horde: a scalable real-time architecture for learning knowledge from
  unsupervised sensorimotor interaction.
\newblock In \emph{10th International Conference on Autonomous Agents and
  Multiagent Systems {(AAMAS} 2011), Taipei, Taiwan, May 2-6, 2011, Volume
  1-3}, pages 761--768, 2011.
\newblock URL
  \url{http://portal.acm.org/citation.cfm?id=2031726\&CFID=54178199\&CFTOKEN=61392764}.

\bibitem[Tamar et~al.(2013)Tamar, Castro, and Mannor]{DBLP:conf/icml/TamarCM13}
Aviv Tamar, Dotan~Di Castro, and Shie Mannor.
\newblock Temporal difference methods for the variance of the reward to go.
\newblock In \emph{Proceedings of the 30th International Conference on Machine
  Learning, {ICML} 2013, Atlanta, GA, USA, 16-21 June 2013}, pages 495--503,
  2013.
\newblock URL \url{http://jmlr.org/proceedings/papers/v28/tamar13.html}.

\bibitem[White(1982)]{WHITE1982639}
D.J White.
\newblock Multi-objective infinite-horizon discounted markov decision
  processes.
\newblock \emph{Journal of Mathematical Analysis and Applications}, 89\penalty0
  (2):\penalty0 639 -- 647, 1982.
\newblock ISSN 0022-247X.
\newblock \doi{https://doi.org/10.1016/0022-247X(82)90122-6}.
\newblock URL
  \url{http://www.sciencedirect.com/science/article/pii/0022247X82901226}.

\bibitem[Yu et~al.(1998)Yu, Lin, and Yan]{YU1998193}
Stella~X Yu, Yuanlie Lin, and Pingfan Yan.
\newblock Optimization models for the first arrival target distribution
  function in discrete time.
\newblock \emph{Journal of Mathematical Analysis and Applications},
  225\penalty0 (1):\penalty0 193 -- 223, 1998.
\newblock ISSN 0022-247X.
\newblock \doi{https://doi.org/10.1006/jmaa.1998.6015}.
\newblock URL
  \url{http://www.sciencedirect.com/science/article/pii/S0022247X98960152}.

\end{thebibliography}

\end{document}